# Greedy Deep Dictionary Learning

Snigdha Tariyal, Angshul Majumdar, *Member IEEE*, Richa Singh, *Senior Member IEEE*, and Mayank Vatsa, *Senior Member, IEEE*

*Abstract*—**In this work we propose a new deep learning tool – deep dictionary learning. Multi-level dictionaries are learnt in a greedy fashion – one layer at a time. This requires solving a simple (shallow) dictionary learning problem; the solution to this is well known. We apply the proposed technique on some benchmark deep learning datasets. We compare our results with other deep learning tools like stacked autoencoder and deep belief network; and state-of-the-art supervised dictionary learning tools like discriminative K-SVD and label consistent K-SVD. Our method yields better results than all.**

## I. Introduction

IN recent years there has been a lot of interest in dictionary learning. However the concept of dictionary learning has been around for much longer. Its application in vision [1] and information retrieval [2] dates back to the late 90's. In those days, the term 'dictionary learning' had not been coined; researchers were using the term 'matrix factorization'. The goal was to learn an empirical basis from the data. It basically required decomposing the data matrix to a basis / dictionary matrix and a feature matrix – hence the name 'matrix factorization'.

The current popularity of dictionary learning owes to K-SVD [3, 4]. K-SVD is an algorithm to decompose a matrix (training data) into a dense basis and sparse coefficients. However the concept of such a dense-sparse decomposition predates K-SVD [5]. Since the advent of K-SVD in 2006, there have been a plethora of work on this topic. Dictionary learning can be used both for unsupervised problems (mainly inverse problems in image processing) as well as for problems arising in supervised feature extraction.

Dictionary learning has been used in virtually all inverse problems arising in image processing starting from simple image [6, 7] and video [8] denoising, image inpainting [9], to more complex problems like color image restoration [10], inverse half toning [11] and even medical image reconstruction [12, 13]. Solving inverse problems is not the goal of this work; we are more interested in dictionary learning from the perspective of machine learning. We briefly discussed [6-13] for the sake of completeness.

Mathematical transforms like DCT, wavelet, curvelet, Gabor etc. have been widely used in image classification problems [14-16]. These techniques used these transforms as a sparsifying step followed by statistical feature extraction methods like PCA or LDA before feeding the features to a classifier. Just as dictionary learning is replacing such fixed transforms (wavelet, DCT, curvelet etc.) in signal processing problems, it is also replacing them in feature extraction scenarios. Dictionary learning gives researchers the opportunity to design dictionaries to yield not only sparse representation (like curvelet, wavelet, DCT etc.) but also discriminative information.

Initial techniques proposed naïve approaches which learnt specific dictionaries for each class [17-19]. Later approaches incorporated discriminative penalties into the dictionary learning framework. One such technique is to include softmax discriminative cost function [20-22]; other discriminative penalties include Fisher discrimination criterion [23], linear predictive classification error [24, 25] and hinge loss function [26, 27]. In [28, 29] discrimination is introduced by forcing the learned features to map to corresponding class labels.

All prior studies on dictionary learning (DL) are 'shallow' learning models just like a restricted boltzman machine (RBM) [30] and autoencoder (AE) [31]. DL, RBM and AE – all fall under the broader topic of representation learning. In DL, the cost function is Euclidean distance between the data and the representation given the learned basis; for RBM it is Boltzman energy; in AE, the cost is the Euclidean reconstruction error between the data and the decoded representation / features.

Almost at the same time, when dictionary learning started gaining popularity, researchers in machine learning observed that better (more abstract and compact) representation can be achieved by going deeper. Deep Belief Network (DBN) is formed by stacking one RBM after the other [32, 33]. Similarly stacked autoencoder (SAE) were created by one AE inside the other [34, 35].

Following the success of DBN and SAE, we propose to learn multi-level deep dictionaries. This is the first work on deep dictionary learning. The rest of the paper will be organized into several sections….

## II. Literature Review

We will briefly review prior studies on dictionary learning, stacked autoencoders and deep Boltzmann machines.

*A. Dictionary Learning*

Early studies in dictionary learning wanted to learn a basis for representation. There were no constraints on the dictionary atoms or on the loading coefficients. The method of optimal directions [36] was used to learn the basis:

$$\min_{D,Z} \|X - DZ\|_F^2 \quad (1)$$

Here $X$ is the training data, $D$ is the dictionary to be learnt and $Z$ consists of the loading coefficients

For problems in sparse representation, the objective is to learn a basis that can represent the samples in a sparse fashion, i.e. $Z$ needs to be sparse. The KSVD [3, 4] is the most well known technique for solving this problem. Fundamentally it solves a problem of the form:

$$\min_{D,Z} \|X - DZ\|_F^2 \text{ such that } \|Z\|_0 \leq \tau \quad (2)$$

KSVD proceeds in two stages. In the first stage it learns the dictionary and in the next stage it uses the learned dictionary to sparsely represent the data. Solving the $l_0$-norm minimization problem is NP hard [37]. KSVD employs the greedy (sub-optimal) orthogonal matching pursuit (OMP) [38] to solve the $l_0$-norm minimization problem approximately. In the dictionary learning stage, KSVD proposes an efficient technique to estimate the atoms one at a time using a rank one update. The major disadvantage of KSVD is that it is a relatively slow technique owing to its requirement of computing the SVD (singular value decomposition) in every iteration. There are other efficient optimization based approaches for dictionary learning [39, 40] – these learn the full dictionary instead of updating the atoms separately.

The dictionary learning formulation in (2) is unsupervised. As mentioned before there is a large volume of work on supervised dictionary learning problems. We will briefly discuss the major ones here. The first work on Sparse Representation based Classification (SRC) [41] was not much of a dictionary learning technique, but was a simple dictionary design problem where all the training samples are concatenated in a large dictionary. The assumption is that the training samples for a basis for any new test sample belonging to the correct class. Their proposed model is:

$$x = Xa \quad (3)$$

where $x$ is the test sample and $X$ is dictionary consisting of all the training samples.

It is assumed in [41] that since the correct class only represents x, the vector $a$ is going to be sparse. Based on this assumption they solved a using some sparse recovery technique. Once $a$ is obtained, the problem is to classify $x$. This is achieved by computing the error between the test image and its representation from each class $c$ obtained by $X^c a^c$. where c denotes the $c^{th}$ class. The test sample is simply assigned to the class having the lowest error.

Several improvements to the basic SRC formulation was proposed in [42-44]. In [42, 43] it was proposed that since $a$ has a known class structure, one can improve upon the basic sparse classification approach by incorporating group-sparsity. In [44] a non-linear extension to the SRC was proposed. Later works handled the non-linear extension in a smarter fashion using the kernel trick [45-47].

The SRC does exactly fit into the dictionary learning paradigm. However [48] proposed a simple extension of SRC – instead of using raw training samples as the basis, they learnt a separate basis for each class and used these dictionaries for classification. This approach is naïve; there is no guarantee that dictionaries from different classes would not be similar. In [49] this issue is corrected. Here an additional incoherency penalty on the dictionaries. This penalty assures that the dictionaries from different classes look different from each other. The formulation is given as:

$$\min_{D_i, Z_i} \sum_{i=1}^{C} \left\{ \|X_i - DZ_i\|_F^2 + \lambda \|Z_i\|_1 \right\} + \eta \sum_{i \neq j} \|D_i^T D_j\|_F^2 \quad (4)$$

Unfortunately this formulation does not improve the overall results too much. It learns dictionaries that look different from each other but does not produce features that are distinctive; i.e. the feature generated for the test sample from dictionaries of all classes looked more or less the same.

The aforesaid issue was rectified in [50]; it combined two concepts. The first one is the discrimination of the learned features and the second one is the discrimination of the class specific dictionaries. The second criteria demands that the features from a particular class will reconstruct the samples of the same class accurately; however it will not represent samples of the other classes. This idea is formulated as follows:

$$C(X_i D, Z_i) = \|X_i - DZ_i\|_F^2 + \|X_i - D_i Z_i^i\|_F^2 + \sum_{i \neq j} \|D_j Z_i^j\|_F^2 \quad (5)$$

Here $D = [D_1 | ... | D_c | ... | D_C]$ is the augmented dictionary and $D_c$ are the class specific dictionaries, $X_i$ are the training samples for the $i^{th}$ class, $Z_i$ is the representaion over all the dictionaries. According to their assumption, only the portion of $Z_i$ pertaining to the correct class should represent the data well - this leads to the second term in the expression; the other dictionaries should not represent the data well hence the third term.

So far, we have discussed about the discriminative dictionaries. As mentioned before, [50] has a second term that discriminates among the learned features. This term arises from the Fisher Discriminant Analysis - it tries to increase the covariance between the classes and decrease covariance within the class. This is represented by:

$$f(Z) = tr(S_W) - tr(S_B) + \eta \|Z\|_F^2 \quad (6)$$

where $S_W = \sum_{c=1}^{C} \sum_i (z_i - \bar{z}_c)(z_i - \bar{z}_c)^T$ and

$S_W = \sum_{c=1}^{C} (\bar{z}_c - \bar{z})(\bar{z}_c - \bar{z})^T$; the regularization helps in stabilizing the solution.

The complete formulation given in [50] is as follows:

$$\min_{D,Z} C(XD, Z) + \lambda_1 \|Z\|_1 + \lambda_2 f(Z) \quad (7)$$

The label consistent KSVD is one of the more recent techniques for learning discriminative sparse representation. It is simple to understand and implement; it showed good results for face recognition [28, 29]. The first technique called Discriminative K-SVD [28] or LC-KSVD1 [29]; it proposes an optimization problem of the following form:

$$\min_{D,Z,A} \|X - DZ\|_F^2 + \lambda_1 \|D\|_F^2 + \lambda_2 \|Z\|_1 + \lambda_3 \|Q - AZ\|_F^2 \quad (8)$$

Here $Q$ is the label of the training samples, it is a canonical basis with a one for the correct class and zeroes elsewhere. $A$ is a parameter of the linear classifier.

In [29] a second formulation is proposed that adds another term to penalize classification error. The LC-KSVD2 formulation is as follows:

$$\min_{D,Z,A,W} \|X - DZ\|_F^2 + \lambda_1 \|D\|_F^2 + \lambda_2 \|Z\|_1 \\ + \lambda_3 \|Q - AZ\|_F^2 + \lambda_4 \|H - WZ\|_F^2 \quad (9)$$

$H$ is a 'discriminative' sparse code corresponding to an input signal sample, if the nonzero values of $H_i$ occur at those indices where the training sample $X_i$ and the dictionary item $d_k$ share the same label. Basically this formulation imposes labels not only on the sparse coefficient vectors $Z_i$'s but also on the dictionary atoms.

During training, the LC-KSVD learns a discriminative dictionary $D$. The dictionary $D$ and the classification weights $A$ need to be normalized. When there is a new test sample, the sparse coefficients for the same are learnt using normalized dictionary using $l_1$-minimization:

$$z_{test} = \min_z \|x_{test} - Dz\|_2^2 + \lambda \|z\|_1 \quad (10)$$

Once the sparse representation of the test sample is obtained, the classification task is straightforward – the label of the test sample is assigned as:

$$j = \arg\max_j (Az_{test}) \quad (11)$$

*B. Deep Boltzman Machine*

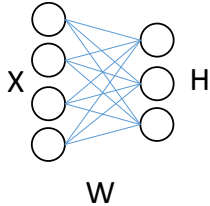

Fig. 1. Restricted Boltzman Machine

Restricted Boltzmann Machines are undirected models that uses stochastic hidden units to model the distribution over the stochastic visible units. The hidden layer is symmetrically connected with the visible unit and the architecture is "restricted" as there are no connections between units of the same layer. Traditionally, RBMs are used to model the distribution of the input data p(x).

The schematic diagram of RBM is shown in Fig. 1. The objective is to learn the network weights (W) and the representation (H). This is achieved by optimizing the Boltzman cost function given by:

$$p(W,H) = e^{-E(W,H)} \quad (12)$$

Where, $E(W,H) = -H^T W X$ including the bias terms.

The conditional distributions are given by (assuming independence) –

$$p(X|H) = \prod p(x|h)$$
$$p(H|X) = \prod p(h|x)$$

Assuming binary input variable, the probability that a node will be active can be given as follows,

$$p(x=1|h) = sigm(W^T h)$$
$$p(h=1|x) = sigm(Wx)$$

Computing the exact gradient of this loss function is almost intractable. However, there is a stochastic approximation to approximate the gradient termed as contrastive divergence gradient. A sequence of Gibbs sampling based reconstruction, produces an approximation of the expectation of joint energy distribution, using which the gradient can be computed.

Usually RBM is unsupervised, but there are studies which trained discriminative RBMs by utilizing the class labels [51]. There are also RBMs which are sparse [52]; the sparsity is controlled by the firing the hidden units only if they are over some threshold. Supervision can also be achieved using sparse RBMs by extending it to have similar sparsity structure within the group / class [53].

Deep Boltzmann Machines (DBM) [54] is an extension of RBM by stacking multiple hidden layers on top of each other (Fig. 2). DBM is an undirected learning model and thus it is different from the other stacked network architectures that each layer receives feedback from both the top-down and bottom-up layer signals. This feedback mechanism helps in managing uncertainty in learning models. While the traditional RBM can model logistic units, a Gaussian-Bernoulli RBM [55] can be used as well with real valued visible units.

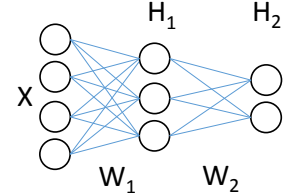

Fig. 2. Deep Boltzman Machine

*C. Stacked Autoencoder*

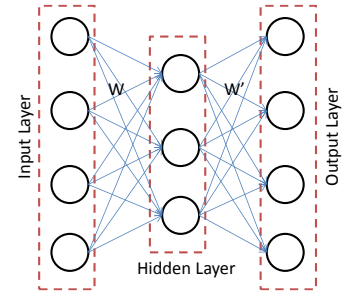

Fig. 3. Single Layer Autoencoder

An autoencoder consists (as seen in Fig. 3) of two parts – the encoder maps the input to a latent representation, and the decoder maps the latent representation back to the data. For a given input vector (including the bias term) *x*, the latent space is expressed as:

$$h = Wx \qquad (13)$$

Here the rows of *W* are the link weights from all the input nodes to the corresponding latent node. Usually a non-linear activation function is used at the output of the hidden nodes leading to:

$$h = \phi(Wx) \qquad (14)$$

The sigmoid function is popular; other non-linear activation functions (like tanh) can be used as well. Rectifier units and large neural networks employ linear activation functions (identity) – this considerably speeds up training.

The decoder portion reverse maps the latent variables to the data space.

$$x = W'\phi(Wx) \qquad (15)$$

Since the data space is assumed to be the space of real numbers, there is no sigmoidal function here.

During training the problem is to learn the encoding and decoding weights – *W* and *W'*. These are learnt by minimizing the Euclidean cost:

$$\arg\min_{W,W'} \|X - W'\phi(WX)\|_F^2 \qquad (16)$$

Here $X = [x_1 |...| x_N]$ consists all the training sampled stacked as columns. The problem (16) is clearly non-convex, but is smooth and hence can be solved by gradient descent techniques; the activation function needs to be smooth and continuously differentiable.

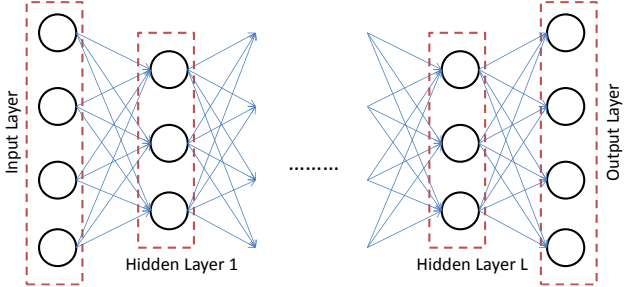

Fig. 4. Stacked Autoencoder

There are several extensions to the basic autoencoder architecture. Stacked autoencoders have multiple hidden layers – one inside the other (see Fig. 4). The corresponding cost function is expressed as follows:

$$\arg\min_{W_1...W_{L-1},W'_1...W'_L} \|X - g \circ \qquad (17)$$

where $g = W_1'\phi(W_2'...W_L'(f(X)))$ and

$f = \phi(W_{L-1}\phi(W_{L-2}...\phi(W_1 X)))$

Solving the complete problem (17) is computationally challenging. Also learning so many parameters (network weights) lead to over-fitting. To address both these issues, the weights are usually learned in a greedy fashion layer by layer [32, 34].

Stacked denoising autoencoder [35] is a variant of the basic autoencoder where the input consists of noisy samples and the output consists of clean samples. Here the encoder and decoder are learnt to denoise noisy input samples.

Another variation for the basic autoencoder is to regularize it, i.e.

$$\arg\min_{(W)s} \|X - g \circ \qquad + R(W, X) \qquad (18)$$

The regularization can be a simple Tikhonov regularization – however that is not used in practice. It can be a sparsity promoting term [56, 57] or a weight decay term (Frobenius norm of the Jacobian) as used in the contractive autoencoder [58]. The regularization term is usually chosen so that they are differentiable and hence minimizable using gradient descent techniques.

### III. DEEP DICTIONARY LEARNING

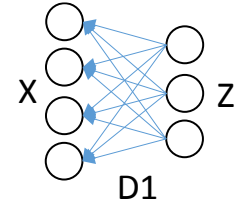

Fig. 5. Schematic Diagram for Dictionary Learning

In this section we describe the main contribution of this work. A single / shallow level of dictionary learning yields a latent representation of data and the dictionary atoms. Here we propose to learn deeper latent representation of data by learning multi-level dictionaries. The idea of learning deeper levels of dictionaries stems from the recent success of deep learning in various areas of machine learning.

The schematic diagram for dictionary learning is shown in Fig. 5. *X* is the data, *D* is the dictionary and *Z* is the feature / representation of X in D. Dictionary learning follows a synthesis framework, i.e. the dictionary is learnt such that the features synthesize the data along with the dictionary.

$$X = DZ \qquad (19)$$

There is also analysis K-SVD, but it cannot be used for feature extraction, it can only produce a 'clean' version of the data and hence is only suitable for inverse problems.

In this work, we propose to extend the shallow (Fig. 3) dictionary learning into multiple layers – leading to deep dictionary learning (Fig. 6).

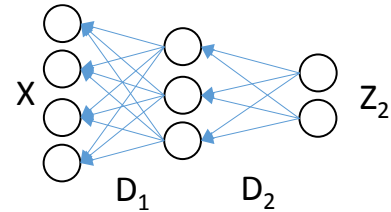

Fig. 6. Schematic Diagram for Deep Dictionary Learning

Mathematically, the representation at the second layer is represented as:

$$X = D_1 D_2 Z_2 \quad (20)$$

Learning the two dictionaries along with the deepest level features is a hard problem for two reasons:
1) Dictionary learning (19) is a bi-linear (hence non-convex) problem. Learning multiple layers of dictionaries along with the features makes the problem even more difficult to solve. Only recently, studies have proven some convergence guarantees for single level dictionary learning [59-63]. These proofs would be very hard to replicate for multiple layers.
2) Moreover, the number of parameters required to be solved increases when multiple layers are dictionaries are learnt simultaneously. With limited training data, this could lead to over-fitting.

Here we propose to learn the dictionaries in a greedy fashion. This is in sync with other deep learning techniques [32-34]. Moreover, layer-wise learning will guarantee the convergence at each layer. The diagram illustrating layer-wise learning is shown in Fig. 5.

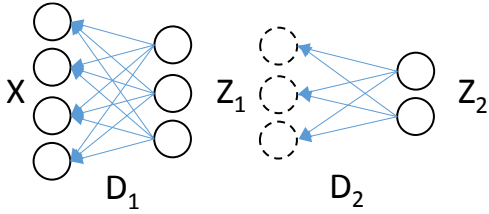

Fig. 7. Greedy Layer-wise Learning

In a greedy fashion, we start with the first layer, i.e. we solve for $D_1$ and $Z_1$ from –
$$X = D_1 Z_1 \quad (21)$$
The features from the first layer ($Z_1$) acts as input to the second layer. Therefore the second layer learns the weights from –
$$Z_1 = D_2 Z_2 \quad (22)$$
The learning can be either dense or sparse, i.e. the features / representation can be dense or sparse. For dense features, the learning is simple and is given by (23)
$$\min_{D,Z} \|X - DZ\|_2^2 \quad (23)$$
Optimality of solving (23) by alternating minimization has been proven in [56]. Therefore we follow the same. The dictionary D and the basis Z is learnt by:
$$Z_k \leftarrow \min_Z \|X - D_{k-1} Z\|_2^2 \quad (24a)$$
$$D_k \leftarrow \min_D \|X - DZ_k\|_2^2 \quad (24b)$$
This is simply the method of optimal directions [36]. Both (24a) and (24b) are simple least square problems having closed form solutions.

For learning sparse features, one just needs to regularize (23) by an $l_1$-norm on the features. This is given by:
$$\min_{D,Z} \|X - DZ\|_2^2 + \lambda \|Z\|_1 \quad (25)$$
This too is solved using alternating minimization.

$$Z_k \leftarrow \min_Z \|X - D_{k-1} Z\|_2^2 + \lambda \|Z\|_1 \quad (26a)$$
$$D_k \leftarrow \min_D \|X - DZ_k\|_2^2 \quad (26b)$$

As before, solving (26b) is simple. It is a least square problem having a closed form solution. The solution to (26a) although not analytic, is well known in signal processing and machine learning literature. It can solved using the Iterative Soft Thresholding Algorithm (ISTA) [64]. In every iteration, the steps for ISTA are:

$$B = Z + \frac{1}{\alpha} D_{k-1}^T (X - D_{k-1} Z)$$

$$Z \leftarrow signum(B) \max\left(0, |B| - \frac{\lambda}{2\alpha}\right)$$

In this work, we have used dense dictionary learning for all layers till the penultimate layer and sparse dictionary learning only in the final layer, i.e. for the two layer problem, the first layer ($D_1$, $Z_1$) would be dense and the second layer ($D_2$, $Z_2$) would be sparse.

It must be noted that the two dictionaries cannot be collapsed into a single one. This is because the learning process is non-linear. For example, if the dimensionality of the sample is m and the first dictionary is of size m x $n_1$ and the second one is $n_1$ x $n_2$, it is not possible to learn a single dictionary of size m x $n_2$ and expect the same results as a two-stage dictionary.

### A. Connection with RBM

RBM is an undirectional graph, whereas dictionary learning is unidirectional. This is evident from figures 1 and 5. In both cases, the task is to learn the network weights / atoms and the representation given the data. They differ from each other in the cost functions used. For RBM it is the Boltzmann function. Here one tries to learn the network weight and the output features such that the similarity between the projected data (at the input) and the features is maximized.

In dictionary learning, the cost function is different – instead of maximizing similarity, we minimize the Euclidean distance between the data (X) and the synthesis (DZ). RBM has a stochastic formulation; dictionary learning is deterministic.

RBMs can be formulated for features having values between 0 and 1. If the values are outside this range, they need to be normalized. In many cases, the normalization does not affect the performance, but there can be scenarios where it suppresses important information. Dictionary learning can work both on real and complex inputs.

### B. Connection with Autoencoder

We mentioned before that dictionary learning is predominantly modeled as a synthesis problem, i.e. the dictionary and the features are learnt such that they can synthesize the data. It is expressed as: $X = D_S Z$ where X is the data, $D_S$ is the learnt synthesis dictionary and Z are the sparse coefficients.

Usually one promotes sparsity in the features and the learning requires minimizing the following,

$$\|X - D_S Z\|_F^2 + \lambda \|Z\|_1 \quad (26)$$

This is the so called synthesis prior formulation where the task is to find a dictionary that will synthesize / generate signals from sparse features. There is an alternate co-sparse analysis prior dictionary learning paradigm [65] where the goal is to learn a dictionary such that when it is applied on the data the resulting coefficient is sparse. The model is $D_A \hat{X} = Z$. The corresponding learning problem is framed minimizing:

$$\|X - \hat{X}\|_F^2 + \lambda \|D_A \hat{X}\|_1 \quad (27)$$

If we combine analysis and synthesis, using $\hat{X} = D_S Z, D_A \hat{X} = Z$ and impute it in (27) we get –

$$\|X - D_S D_A \hat{X}\|_F^2 + \lambda \|D_A \hat{X}\|_1 \quad (28)$$

This is the expression of a sparse denoising autoencoder [54] with linear activation at the hidden layer. If we drop the sparsity term, it becomes –

$$\|X - D_S D_A \hat{X}\|_F^2 \quad (29)$$

This formulation is similar to a denoising autoencoder with linear activation.

We can express autoencoder in the lingo of dictionary learning – autoencoder is a model that learnt the analysis and the synthesis dictionaries. To the best of our knowledge, this is the first work which shows the architectural similarity between autoencoders and dictionary learning.

IV. EXPERIMENTAL EVALUATION

A. Datasets

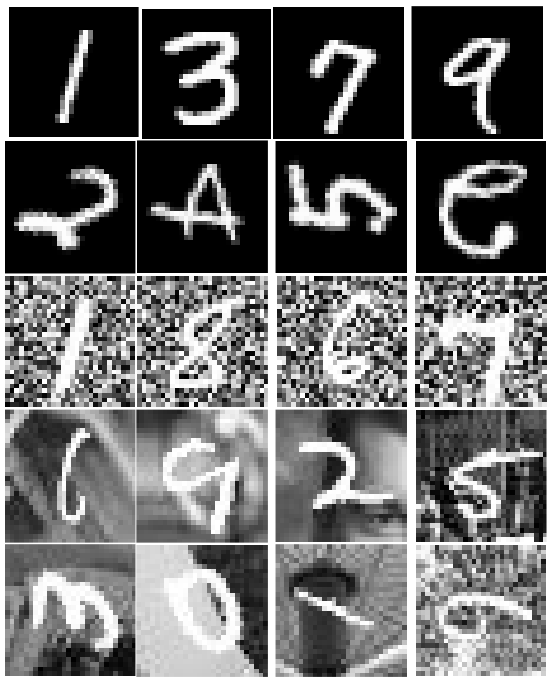

Fig. 8. Top to bottom. basic, basic-rot, bg-rand, bg-img, bg-img-rot

We carried our experiments on several benchmarks datasets. The first one is MNIST dataset which consists of 28x28 images of handwritten digits ranging from 0 to 9. The dataset has 60,000 images for training and 10,000 images for testing. No preprocessing has been done on this dataset.

The other datasets are variations of MNIST, which are more challenging primarily because they have fewer training samples (10,000) and larger number of test samples (50,000).
1. basic (smaller subset of MNIST)
2. basic-rot (smaller subset with random rotations)
3. bg-rand (smaller subset with uniformly distributed noise in background)
4. bg-img (smaller subset with random image background)
5. bg-img-rot (smaller subset with random image background plus rotation)

Samples for each of the datasets are shown in Fig. 8.

B. Deep vs Shallow Dictionary Learning

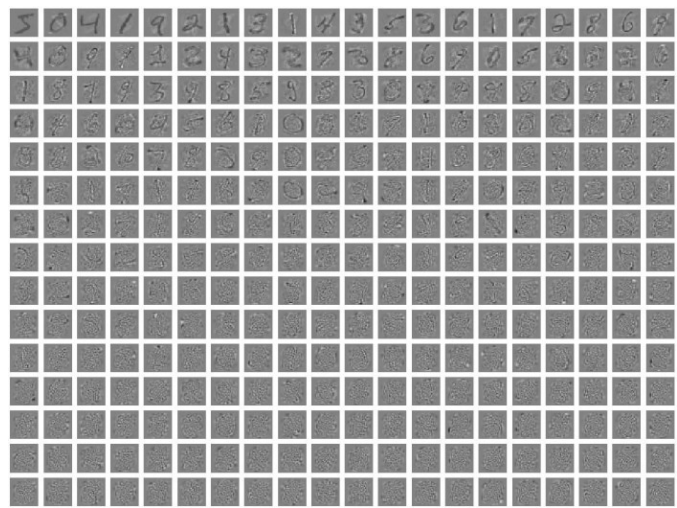

Fig. 9. First level dictionary for MNIST

In the first set of results, we show that the multi-level dictionaries cannot be collapsed into a single one and expected to perform the same. We carried out experiments on the MNIST and its variations. In the first case, the number of basis in the multi-level dictionaries are: 300-15-50. In the second case, we learn a shallow dictionary with 50 atoms. The results from these two would be the same, if the multi-level dictionaries would be collapsible.

We want to show that the representation learnt from a single level of dictionary and multi-level dictionary are different. To showcase this, we show classification results with a simple K Nearest Neighbour (K=1). The classification accuracies are shown in Table 1.

We use a deterministic initialization for dictionary learning. Usually the dictionary atoms are initialized by randomly choosing samples from the training set – but this leads to variability in results. In this work we propose a deterministic initialization based on QR decomposition. Orthogonal vectors from Q (in order) are used to initialize the dictionary.

TABLE I
DEEP VS SHALLOW

| Dataset | Deep (300-15-50) | Shallow (50) |
|---|---|---|
| MNIST | 97.75 | 97.35 |
| basic | 95.80 | 95.02 |
| basic-rot | 87.00 | 84.19 |
| bg-rand | 89.35 | 87.19 |
| bg-img | 81.00 | 78.86 |
| bg-img-rot | 57.77 | 54.40 |

The discrepancy between multi-level dictionary learning and single level dictionary learning is evident in Table 1. If the learning was linear, it would be possible to collapse multiple dictionaries into one; but dictionary learning is inherently non-linear. Hence it is not possible to learn a single layer of dictionary in place of multiple levels and expect the same output.

### C. Comparison with other Deep Learning Approaches

We compared our results with a stacked autoencoder (SAE) and deep belief network (DBN). The implementation for these have been obtained from [66] and [67] respectively. Both SAE and DBN has a three layer architecture. The number of nodes is halved in every subsequent layer. This is a standard approach; we tried other configurations but could not improve upon this.

We want to compare the representation capability of our proposed technique vis-à-vis other deep learning methods. The results for K Nearest Neighbour (KNN) and Support Vector Machine (SVM) are shown in Tables 2 and 3.

TABLE II
COMPARISON WITH KNN (K=1) CLASSIFICATION

| Dataset | DDL | DBN | SAE |
|---|---|---|---|
| MNIST | **97.75** | 97.05 | 97.33 |
| basic | **95.8** | 95.37 | 95.25 |
| basic-rot | **87.00** | 84.71 | 84.83 |
| bg-rand | **89.35** | 77.16 | 86.42 |
| bg-img | 81.00 | **86.36** | 77.16 |
| bg-img-rot | **57.77** | 50.47 | 52.21 |

TABLE III
COMPARISON WITH SVM CLASSIFICATION

| Dataset | DDL | DBN | SAE |
|---|---|---|---|
| MNIST | **98.64** | 98.53 | 98.5 |
| basic | 97.284 | 88.44 | **97.4** |
| basic-rot | **90.344** | 76.59 | 79.83 |
| bg-rand | **92.38** | 78.59 | 85.34 |
| bg-img | **86.17** | 75.22 | 74.99 |
| bg-img-rot | **63.85** | 48.53 | 49.14 |

We find that apart from one case each in Tables 1 and 2, our proposed method yields better results than DBN and SAE. For KNN, our results are slightly better, but for SVM we are doing considerably better, especially for the more difficult datasets.

We have compared our technique with state-of-the-art dictionary learning techniques like D-KSVD [28] and LC-KSVD [29]. These were tuned to yield the best possible results. Comparison is also done with stacked denoising autoencoder (SDAE) and deep belief network (DBN) fine tuned with softmax classifier. We did not run these experiments; these results are copied from [35].

TABLE IV
COMPARISON WITH OTHER TECHNIQUES

| Dataset | DDL-SVM | LC-KSVD | D-KSVD | DBN-SM* | SDAE-SM* |
|---|---|---|---|---|---|
| MNIST | 98.64 | 93.30 | 93.6 | **98.76** | 98.72 |
| basic | **97.28** | 92.70 | 92.20 | 96.89 | 97.16 |
| basic-rot | 90.34 | 48.66 | 50.01 | 89.7 | **90.47** |
| bg-rand | 92.38 | 87.70 | 87.70 | **93.27** | 89.7 |
| bg-img | **86.17** | 80.65 | 81.20 | 83.69 | 83.32 |
| bg-img-rot | **63.85** | 75.40 | 75.40 | 52.61 | 56.24 |

*Results from [35]

We find that the proposed deep dictionary learning techniques always yields better results than shallow dictionary learning (LC-KSVD and D-KSVD). In most cases, we can even achieve better accuracy than highly tuned models like DBN and SDAE.

We compare our technique with other deep learning approaches in terms of speed (training time). All the algorithms are run until convergence. SAE, DBN and DDL (proposed) are run until convergence. The machine used is Intel (R) Core(TM) i5 running at 3 GHz; 8 GB RAM, Windows 10 (64 bit) running Matlab 2014a. The run times for all the smaller MNIST variations are approximately the same. So we only report results for the larger MNIST dataset (60K) and the basic (10K) dataset.

TABLE II
TRAINING TIME IN SECONDS

| Dataset | DDL | DBN | SAE |
|---|---|---|---|
| MNIST | **107** | 30071 | 120408 |
| basic | **26** | | |

We see that our proposed deep dictionary learning algorithm is more than 2 orders of magnitude faster than deep belief network and more than 3 orders of magnitude faster than stacked autoencoder. This is a huge saving in training time.

### V. CONCLUSION

In this work we propose the idea of deep dictionary learning, where instead of learning one shallow dictionary – as has been done so far, we learn multiple levels of dictionaries. Learning all the dictionaries makes the problem highly non-convex. Also learning so many parameters (atoms of many dictionaries) is always fraught with the problem of over-fitting. To account for both these issues, we learn the dictionaries in a greedy fashion

– one layer at a time. The representation / feature from one level is used as the input to learn the following level. Thus, the basic unit of deep dictionary learning is a simple shallow dictionary learning algorithm; which is a well known and solved problem.

We compare the new deep learning tool with the existing ones like the stacked autoencoder and deep belief network. We find that our method yields better results on benchmark deep-learning datasets. The main advantage of our method is that it is few orders of magnitude faster than existing deep learning tools like stacked autoencoder and deep belief network.

This is a preliminary work, we will carry out more extensive experimentation in the future. We plan to test the robustness of dictionary learning in the presence of missing data, noise and limited number of training sample.

In the future, we would also like to apply this technique for other practical problems arising, biometrics, vision, speech processing etc. Also there has been a lot of work on supervised dictionary learning; our preliminary formulation is unsupervised. In future, we expect to improve the results even further by incorporating techniques from supervised learning.